\documentclass[letterpaper, 10 pt, conference]{ieeeconf}  

\IEEEoverridecommandlockouts                              

\usepackage{graphicx} 
\usepackage{subcaption} 
\usepackage{float}      
\usepackage{amsmath} 
\usepackage{amsfonts}
\usepackage{algorithm, algpseudocode}
\usepackage{gensymb}
\renewcommand{\theenumi}{(\roman{enumi})}

\title{\LARGE \bf
Ocean Current-Harnessing Stage-Gated MPC: Monotone Cost Shaping and Speed-to-Fly for Energy-Efficient AUV Navigation}

\author{Spyridon Syntakas and Kostas Vlachos
\thanks{}
\thanks{Spyridon Syntakas and Kostas Vlachos are with the Department of
Computer Science and Engineering, University of Ioannina, 45110 Ioannina,
Greece. (Email: {ssyntakas; kostaswl}@cse.uoi.gr)}
}

\begin{document}

\maketitle
\thispagestyle{empty}
\pagestyle{empty}

\begin{abstract}

Autonomous Underwater Vehicles (AUVs) are a highly promising technology for ocean exploration and diverse offshore operations, yet their practical deployment is constrained by energy efficiency and endurance. To address this, we propose Current-Harnessing Stage-Gated MPC, which exploits ocean currents via a per-stage scalar which indicates the "helpfulness" of ocean currents. This scalar is computed along the prediction horizon to gate lightweight cost terms only where the ocean currents truly aids the control goal. The proposed cost terms, that are merged in the objective function, are (i) a Monotone Cost Shaping (MCS) term, a help-gated, non-worsening modification that relaxes along-track position error and provides a bounded translational energy rebate, guaranteeing the shaped objective is never larger than a set baseline, and (ii) a speed-to-fly (STF) cost component that increases the price of thrust and softly matches ground velocity to the ocean current, enabling near zero water-relative “gliding.” All terms are $C^1$ and integrate as a plug-and-play in MPC designs. Extensive simulations with the BlueROV2 model under realistic ocean current fields show that the proposed approach achieves substantially lower energy consumption than conventional predictive control while maintaining comparable arrival times and constraint satisfaction.
\end{abstract}

\section{Introduction}

According to recent literature, less than $70\%$ of the ocean floor is mapped to modern standards \cite{seabed}, under $0.001\%$ of the deep seafloor has been viewed \cite{coverage_estimate} while over $80\%$ of the ocean remains unexplored. Autonomous Underwater Vehicles (AUVs) could change this, but mission range is constrained by energy efficiency. In a vast, current-dominated oceanic environment, endurance is the primary bottleneck. Developing energy-efficient controllers for navigation and coverage would let AUVs undertake far more oceanographic workload while also benefit most other marine applications as well.

Although ocean currents are commonly treated as disturbances to be rejected, recent works increasingly exploit ambient ocean dynamics as a source of propulsive benefit and energy efficiency. Prior work leverages ocean currents primarily through energy-aware path planning. Approaches that include sampling-based methods \cite{RRT_pp}, heuristics \cite{A*}, and evolutionary computing \cite{genetic_algo} all aim to increase energy efficiency via planning. For fully propelled AUVs, current-aware planners steer vehicles into favorable corridors subject to obstacle constraints \cite{fully_propeled}. For gliders, strategies exploit forecasts to boost navigation efficiency \cite{article_lag}. Stream-function formulations synthesize trajectories by overlaying control inputs on the background flow \cite{stream_lines}. Despite these benefits, most solutions are offline or weakly coupled to closed-loop control. The proposed Current-Harnessing Stage-Gated MPC closes this gap by incorporating forecast ocean currents along the prediction horizon.

\begin{figure}[h] 
    \centering
    \includegraphics[width=\columnwidth]{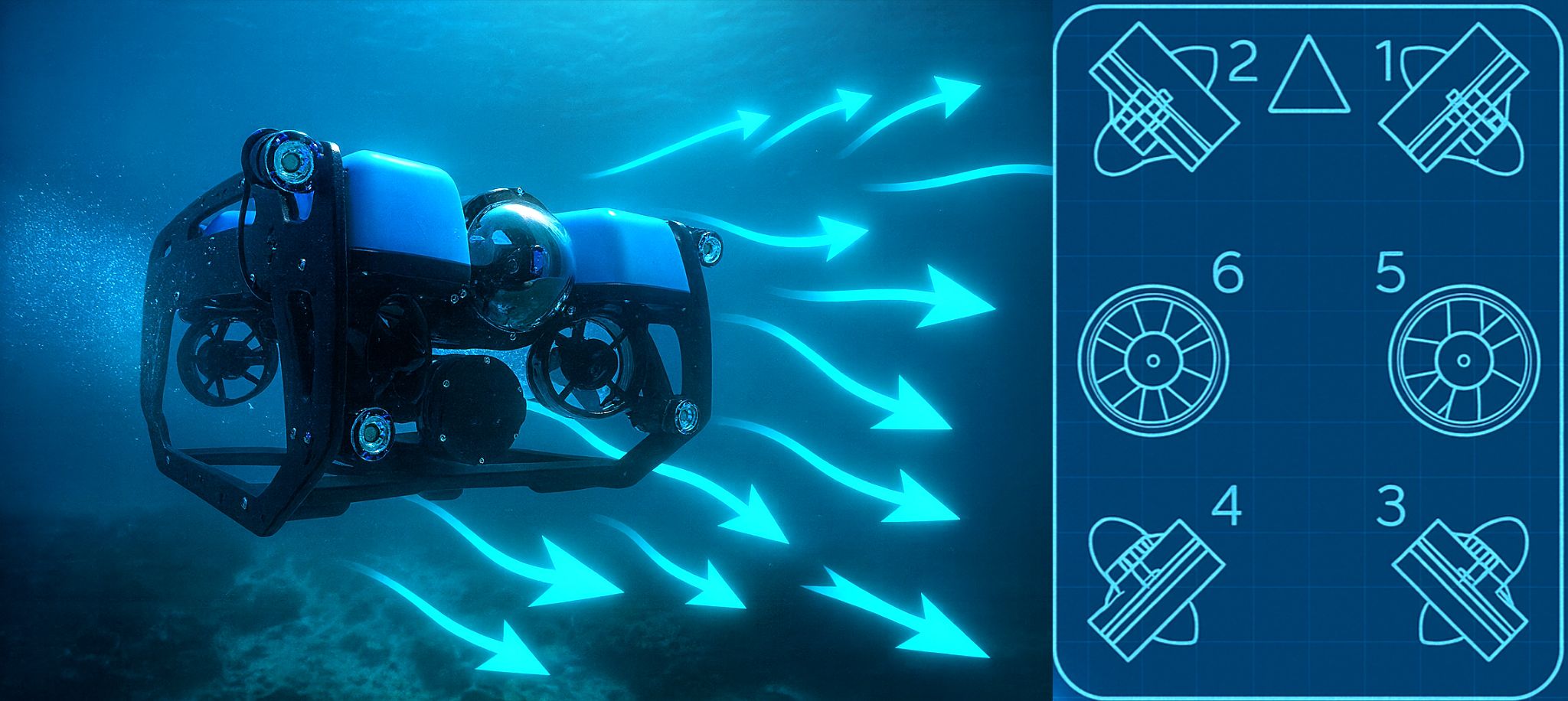} 
    \caption{The BlueROV2 "surfing" the ocean currents for energy-efficiency -  Technical sketch of thruster layout.}
    \label{fig:blue_rov}
\end{figure}

Thrusters are the AUV’s dominant power draw, making energy use both a planning and a control problem. Splitting those tasks is counterintuitive, which makes Nonlinear Model Predictive Control (NMPC) a natural solution because it enforces constraints and offers elegant mission goal expressivity via the cost, letting the Nonlinear Program (NLP) jointly handle planning and ocean dynamics. 

Consequently, many MPC formulations have been explored, including Lyapunov-based MPC for trajectory tracking under thruster saturation \cite{Lyapunov_MPC}, multi-objective MPC via Pontryagin’s maximum principle for path following \cite{MOMPC}, and dual MPC that addresses model uncertainty and ocean current disturbances \cite{dualMPC}. Additional examples include an MPC that predicts wave-field disturbances using linear wave theory \cite{MPC_1} and a disturbance-observer-based MPC demonstrated on the BlueROV2 platform \cite{Disturbance_Observer}. A common thread is treating ocean currents as external disturbances to be rejected, often causing thruster saturation and increased power consumption. 

In contrast to existing methods, the proposed MPC framework explicitly incorporates ocean currents within its cost formulation to achieve energy-efficient navigation. Prior studies include an MPC that integrates ocean currents into the dynamic model of a BlueROV2-type AUV \cite{ICRA_2018}, which demonstrated reduced power consumption but did not explicitly structure the objective to exploit ocean currents, as well as an energy-optimal MPC for BlueROV2 that penalizes power consumption without directly accounting for flow dynamics \cite{energy_optimal}. A complementary research line couples NMPC with A* to generate energy-efficient trajectories \cite{MPC_2}. The present work extends these approaches by embedding ocean current information directly into the cost function, thereby enabling predictive exploitation of ocean currents in trajectory planning and control.

We present a Current-Harnessing Stage-Gated MPC framework for energy-efficient AUV navigation. A per-stage "helpfulness" scalar that combines ocean current–goal alignment and ocean current strength along the prediction horizon, gates two complementary terms used in combination: 
\begin{enumerate}
\item \emph{Monotone Cost Shaping (MCS)}, a help-gated, non-increasing modification that relaxes along-track position error and provides a bounded translational-energy rebate, and 
\item \emph{Speed-to-Fly (STF)}, which increases the marginal cost of thrust and softly aligns ground velocity with the local ocean current to enable gliding when advantageous, taking intuition from para-gliders.
\end{enumerate}
All components are $C^1$ and integrate into state-of-the-art frameworks, i.e., CasADi\cite{casadi}-IPOPT\cite{Ipopt}, with minimal code. The proposed controller consistently reduces energy without meaningfully degrading arrival time or constraint satisfaction and is tested in BlueROV2-based simulations using real ocean current fields from data of the Copernicus Marine Service \cite{Copernicus}.

Namely, the paper makes the following contributions:
\begin{itemize}
\item \textbf{Horizon-wise ocean current incorporation:} At each MPC update, forecast ocean currents are sampled at the \emph{predicted} states across the horizon, enabling the entire prediction to bend toward favorable flow corridors.
\item \textbf{Stage-gated ocean current exploitation:} A per-stage "helpfulness" scalar, embodying ocean current–goal alignment and ocean current strength, gates ocean current exploitation terms, activating them only where the flow is beneficial.
\item \textbf{Monotone Cost Shaping (MCS):} A help-gated, non-increasing modification of the stage cost that (i) relaxes only the along-track position error and (ii) provides a bounded translational-energy rebate, guaranteeing $J_{\text{MCS}}\leq J_{\text{base}} \:\forall\: (x,u)$, where $J$ denotes the cost.
\item \textbf{Speed-to-Fly (STF) integration:} Cost terms that increase the marginal price of thrust and softly align ground velocity with the local ocean current, promoting near–zero water-relative motion when conditions allow.
\item \textbf{Solver-friendly smoothness:} All gates and normalizations are $C^{1}$, improving numerical robustness with gradient-based NLP solvers.
\end{itemize}

\section{Preliminaries}
\subsection{Ocean Currents}\label{sec:ocean_currents}

\begin{figure}[t] 
    \centering
    \includegraphics[width=0.9\columnwidth]{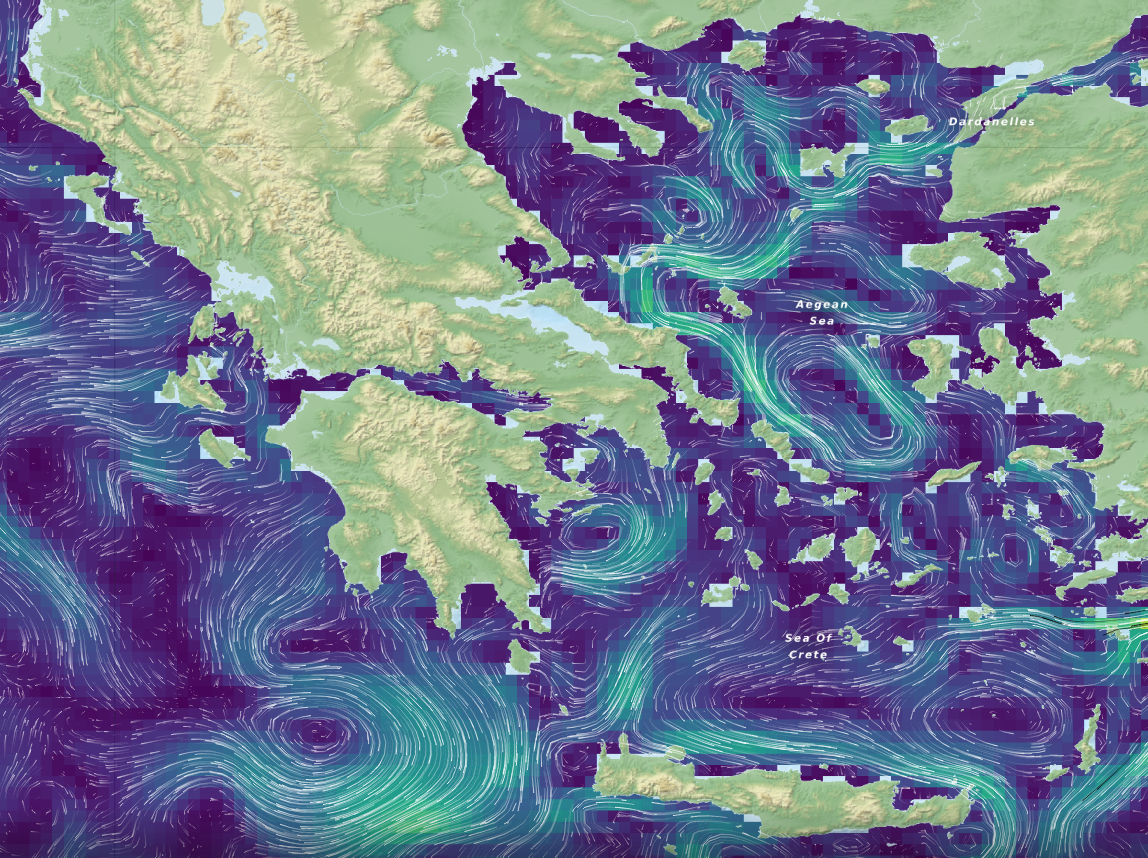} 
    \caption{Oceanic Currents at 50 meters depth in the Aegean and Ionian Sea from the Copernicus Marine Service.}
    \label{fig:currents_my_ocean_pro}
\end{figure}

Before the big data era, ocean currents had to be inferred from limited and often uncertain data. Today, open-access resources provide near real-time oceanic measurements on a global scale. This constitutes a crucial advantage for marine applications. In the case of AUVs, this enables the development of algorithms that leverage ocean currents using observable, near real-time data, thereby enhancing their effective exploitation.

Copernicus Marine Service \cite{Copernicus}, a pillar of the EU Space Program, provides free, science-based and trusted ocean data. While in-situ observations provide limited coverage, especially at high latitudes, the accessibility of Copernicus ocean current data is steadily improving, due to the integration of satellite measurements, advanced modeling, and ongoing data assimilation. The data that are provided for subsurface ocean currents have a spatial resolution of $~1/12^{\circ} (\approx 8 km)$ with 50 vertical levels that range to $5.500 m$ depth. An example from currents in the Aegean
and Ionian Sea in the Mediterranean is depicted in Figure \ref{fig:currents_my_ocean_pro}. 

\begin{figure}[t]
    \centering
    
    \begin{subfigure}[b]{0.6\columnwidth}
        \centering
        \includegraphics[width=\columnwidth]{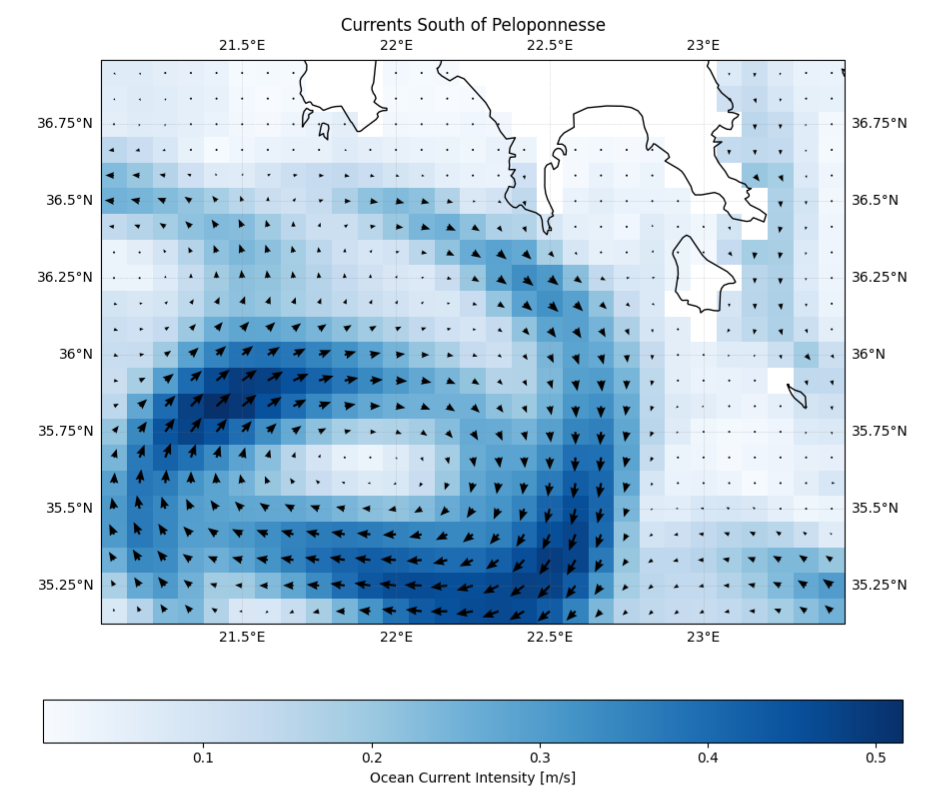}
        \caption{}
        \label{fig:image1}
    \end{subfigure}
    \hfill
    \begin{subfigure}[b]{0.5\columnwidth}
        \centering
        \includegraphics[width=\columnwidth]{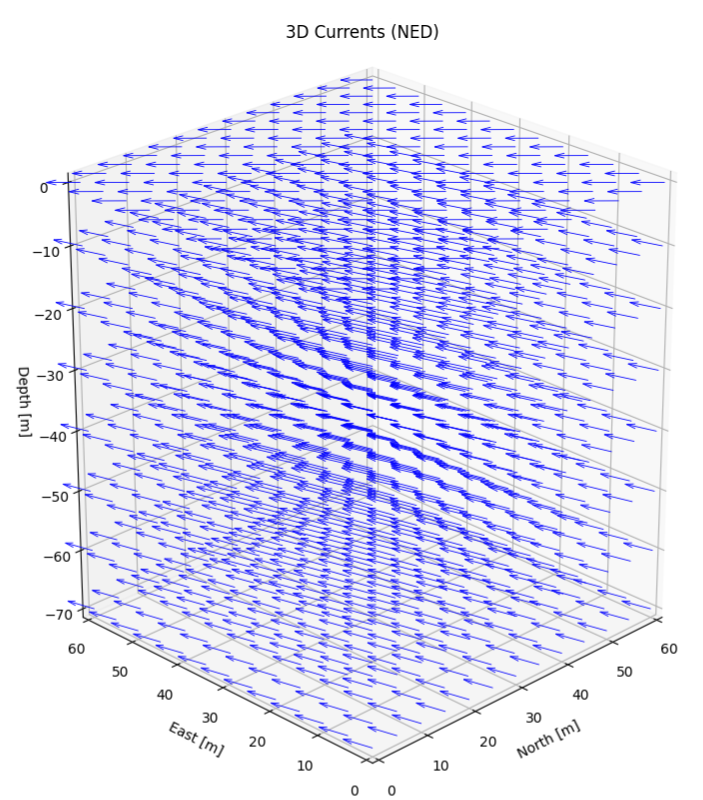}
        \caption{}
        \label{fig:image2}
    \end{subfigure}
    
    \caption{(a) Top view of currents 100m depth south of Peloponnese, Greece. (b) Local 3D current grid in NED.}
    \label{fig:peloponnesse}
\end{figure}

Since ocean currents locally exhibit only minor variations both temporally and in magnitude, careful interpolation of Copernicus data enables the construction of a finely resolved 3D grid describing ocean current dynamics, even for small changes in latitude, longitude, and depth. Figure~\ref{fig:peloponnesse} presents a top view of the ocean currents in a broader region south of the Peloponnese, Greece at a depth of 100 m, within the bounds $lat\in[35^\circ, 37^\circ], \:long\in[21^\circ, 23.5^\circ]$. This region corresponds to a north–south span of $223 km$ and an east–west span of $225 km$, with a spatial ocean current resolution of approximately $9 km$. By applying both vertical and horizontal interpolation, it is possible to zoom into smaller subregions, as illustrated in the same figure, where a $60\times60\times70m$ volume is depicted and referenced in a local NED frame.  These finely resolved grids are employed in the proposed MPC framework, while data can be easily obtained using the Copernicus Marine Services API.
\section{Vehicle Kinematics and Dynamics}\label{sec:kinmatic_dynamics}
The robot of choice, depicted in Figure \ref{fig:blue_rov}, is the BlueROV2 treated as an autonomous underwater vehicle. Although the full 6-DOF dynamics are modeled and simulated, only four DOFs are actively controlled (surge, sway, heave, yaw).  This matches the BlueROV2’s thruster layout and is common practice  for such systems, where roll and pitch are stabilized passively via restoring forces and hydrodynamic damping. In addition compared to other works, that make simplifications, in this work Rigid-Body (RB) terms use absolute velocity, while the Hydrodynamic terms are modeled using the relative velocity. The only simplification that is made and is discussed in the following subsection is that of a quasi-static ocean current field assumption.

\subsection{State Variables}
The BlueROV2 is modeled as a fully nonlinear 6-DOF underwater vehicle, following Fossen's notation \cite{bookFOSSEN}. 
The system state is
\begin{equation}
\mathbf{x} =
\begin{bmatrix}
\boldsymbol{\eta} \\ \boldsymbol{\nu}
\end{bmatrix}
\end{equation}
where $\boldsymbol{\eta}=[x_N,y_E,z_D,\phi,\theta,\psi]^\top$ denotes the pose in the North--East--Down (NED) inertial frame, and $\boldsymbol{\nu}=[u,v,w,p,q,r]^\top$ are the linear and angular velocities in the body-fixed frame.

\subsection{Control Inputs}
As mentioned the BlueROV2 is considered underactuated. Thrusters can generate generalized forces/torque in surge ($X$), sway ($Y$), heave ($Z$), and yaw ($N$). 
The control input vector is therefore $\mathbf{u} =
\begin{bmatrix}
X & Y & Z & N
\end{bmatrix}^\top$
which maps to the 6-DOF generalized vector as
\begin{equation}
\boldsymbol{\tau} =
\begin{bmatrix}
X & Y & Z & 0 & 0 & N
\end{bmatrix}^\top .
\end{equation}

\subsection{Kinematics}
The kinematic equations are writen as
\begin{equation}
\dot{\boldsymbol{\eta}} = J(\boldsymbol{\eta}) \, \boldsymbol{\nu} = 
\begin{bmatrix}
R_{nb}(\phi,\theta,\psi) & 0_{3\times 3} \\
0_{3\times 3} & T(\phi,\theta)
\end{bmatrix}\, \boldsymbol{\nu},
\end{equation}
with $R_{nb}(\phi,\theta,\psi)$ denoting the body-to-NED rotation matrix and 
$T(\phi,\theta)$ the Euler-angle transformation mapping angular velocities 
$\boldsymbol{\omega} = [\,p, q, r\,]^\top$ to Euler angle rates 
$[\,\dot{\phi}, \dot{\theta}, \dot{\psi}\,]^\top$.

Ocean currents are modeled as a steady velocity field $\mathbf{v}_c^n=[u_c^n,v_c^n,0]^\top$ expressed in NED with vertical ocean currents considered negligible.
Its body-fixed representation is
\begin{equation}
\mathbf{v}_c^b = R_{nb}^\top \mathbf{v}_c^n ,
\end{equation}
so the relative velocity is
\begin{equation}
\boldsymbol{\nu}_r = 
\begin{bmatrix}
u - u_c^b & v - v_c^b & w & p & q & r
\end{bmatrix}^\top .
\end{equation}
The translational kinematics become $\dot{\mathbf{x}}_{n}^{pos} = R_{nb}\,\mathbf{v}_b = R_{nb}\left(\mathbf{v}_{\text{rel}} + \mathbf{v}_c^b\right)$,
with $x_k^{\text{pos}} = [x_N,y_E,z_D]\in\mathbb{R}^3$ and $\mathbf{v}_{\text{rel}} = [u - u_c^b, v - v_c^b ,w]\in\mathbb{R}^3$. Thus, the kinematics are governed by the absolute body velocity  $\mathbf{v}_b$, while the hydrodynamic forces are functions of the relative velocity $\mathbf{v}_{\text{rel}}$.

\subsection{Dynamics}
The 6-DOF dynamic model is defined as
\begin{equation}
M\dot{\boldsymbol{\nu}}
+ C_{RB}(\boldsymbol{\nu})\,\boldsymbol{\nu}
+ C_A(\boldsymbol{\nu}_r)\,\boldsymbol{\nu}_r
+ D(\boldsymbol{\nu}_r)\,\boldsymbol{\nu}_r
+ \mathbf{g}_\eta(\boldsymbol{\eta})
= \boldsymbol{\tau}
\end{equation}
with $M = M_{RB} + M_A$.
For detailed analysis of the notation used, refer to Fossen’s standard formulations \cite{bookFOSSEN}.
Epigramatically,
$M_{RB}$ is the rigid-body mass and inertia matrix,
\begin{equation}
M_{RB} =
\begin{bmatrix}
m I_{3} & 0_{3\times 3}\\[4pt]
0_{3\times 3} & I
\end{bmatrix}, \qquad
I = \mathrm{diag}(I_{xx}, I_{yy}, I_{zz}),
\end{equation}
$M_A$ is the added-mass matrix, 
\begin{equation}
M_{A} = \mathrm{diag}\!\big(X_{\dot u},\,Y_{\dot v},\,Z_{\dot w},\,K_{\dot p},\,M_{\dot q},\,N_{\dot r}\big).
\end{equation}
$C_{RB}(\boldsymbol{\nu})$ is the rigid-body Coriolis and centripetal matrix, evaluated with absolute velocity $\boldsymbol{\nu}$,
\begin{equation}
C_{RB}(\boldsymbol{\nu}) =
\begin{bmatrix}
0_{3\times 3} & -\,m S(\mathbf{v}_b)\\[4pt]
-\,m S(\mathbf{v}_b) & -\,S(I \boldsymbol{\omega})
\end{bmatrix},
\end{equation}
where $\mathbf{v}_b = [u,v,w]^\top$, $\boldsymbol{\omega} = [p,q,r]^\top$,
and $S(\cdot)$ is the skew operator.
$C_A(\boldsymbol{\nu}_r)$ is the added-mass Coriolis matrix, evaluated with relative velocity $\boldsymbol{\nu}_r$,
\begin{equation}
C_{A}(\boldsymbol{\nu}_r) =
\begin{bmatrix}
0_{3\times 3} & -\,S(\mathbf{a}_1)\\[4pt]
-\,S(\mathbf{a}_1) & -\,S(\mathbf{a}_2)
\end{bmatrix},
\end{equation}
with
\[
\mathbf{a}_1 =
\begin{bmatrix}
X_{\dot u} u_r\\[2pt]
Y_{\dot v} v_r\\[2pt]
Z_{\dot w} w_r
\end{bmatrix}, \qquad
\mathbf{a}_2 =
\begin{bmatrix}
K_{\dot p} p\\[2pt]
M_{\dot q} q\\[2pt]
N_{\dot r} r
\end{bmatrix},
\]
and $\boldsymbol{\nu}_r = [u_r, v_r, w_r, p, q, r]^\top$ the relative velocity.
$D(\boldsymbol{\nu}_r)$ is the hydrodynamic damping matrix, also evaluated with $\boldsymbol{\nu}_r$,
\begin{equation}
\begin{aligned}
D(\boldsymbol{\nu}_r) = \mathrm{diag}\!\big(
&X_u + X_{|u|}\,|u_r|,\;
 Y_v + Y_{|v|}\,|v_r|,\;
 Z_w + Z_{|w|}\,|w_r|,\\
&K_p + K_{|p|}\,|p|,\;
 M_q + M_{|q|}\,|q|,\;
 N_r + N_{|r|}\,|r|
\big).
\end{aligned}
\end{equation}

$\mathbf{g}_\eta(\boldsymbol{\eta})$ is the restoring force and moment vector from gravity and buoyancy, 

\begin{equation}
\mathbf{g}_\eta(\boldsymbol{\eta}) =
\begin{bmatrix}
(W-B)\,\sin\theta \\[4pt]
-(W-B)\,\cos\theta\,\sin\phi \\[4pt]
-(W-B)\,\cos\theta\,\cos\phi \\[4pt]
-\,m g Z_G \cos\theta \sin\phi \\[4pt]
-\,m g Z_G \sin\theta \\[4pt]
0
\end{bmatrix}.
\end{equation}
while $\boldsymbol{\tau}$ is the generalized force/torque vector, containing the 4-DOF control inputs.

This formulation ensures that rigid-body terms always depend on the absolute vehicle motion, while hydrodynamic loads vanish when the vehicle drifts with the ocean current ($\boldsymbol{\nu}=\mathbf{v}_c$). 
A \emph{steady-current assumption} is adopted ($\dot{\mathbf{v}}_c^n = 0$), so no added-mass feedforward term $M_A \dot{\boldsymbol{\nu}}_c$ is required.

\section{Current-Harnessing Stage-Gated NMPC}
\label{sec:nmpc}

We consider a discrete-time NMPC with sampling period $\Delta t$ and prediction horizon $N$. The state $\mathbf{x}_k\in\mathbb{R}^{12}$ (position, attitude, linear and angular velocities) and control $\mathbf{u}_k\in\mathbb{R}^{4}$ (surge, sway, heave, yaw wrench) evolve under the 6-DoF model with relative-flow dynamics presented in Section \ref{sec:kinmatic_dynamics}
\begin{equation}
\mathbf{x}_{k+1} \;=\; \mathbf{x}_k \;+\; \Delta t\, f\!\big(\mathbf{x}_k,\mathbf{u}_k, \mathbf{v}_{c,k}\big),
\label{eq:dynamics}
\end{equation}
where $\mathbf{v}_{c,k}\in\mathbb{R}^3$ is the ocean current (NED) at stage $k$ and $f\!\big(\mathbf{x}_k,\mathbf{u}_k, \mathbf{v}_{c,k}\big) = \dot{\mathbf{x}} =[\dot{\boldsymbol{\eta}},\dot{\boldsymbol{\nu}}]^T$.
\subsection{Horizon-wise ocean current sampling}
To determine the ocean current velocity over the horizon, at each NMPC update, we form a warm-start state trajectory $\{\mathbf{\tilde x}_k\}_{k=0}^{N}$ (from the previous solution or a steady guess) and sample ocean currents along the \emph{predicted} positions:
\begin{equation}
\mathbf{v}_{c,k} \;=\; \mathcal{V}\!\big(\mathbf{\tilde x}_k^{\text{pos}}\big), 
\qquad \tilde x_k^{\text{pos}}\in\mathbb{R}^3,\;\; \mathbf{v}_{c,k}\in\mathbb{R}^3,
\label{eq:vc-sampling}
\end{equation}
where $\mathcal{V}\!:\mathbb{R}^3\!\to\!\mathbb{R}^3$ maps NED position to the ocean current expressed in NED.
Sampling at predicted states renders the entire horizon \emph{current-aware}, and not just reactive at the present position, in contrast to the literature.

\subsection{"Helpfulness" gate $s_k$}
\begin{figure}[h] 
    \centering
    \includegraphics[width=\columnwidth]{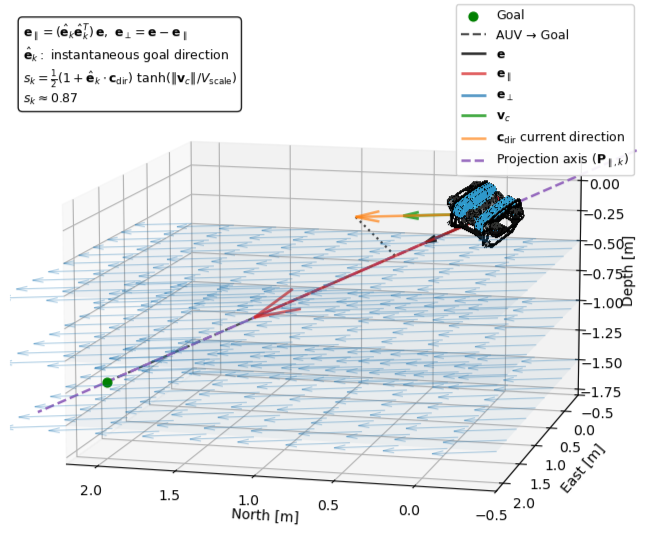} 
    \caption{Geometric intuition of $s_k$ "helpfulness" scalar.}
    \label{fig:sk}
\end{figure}

In order to define the "helpfulness" gate $s_k$, let $x_g$ denote the goal reference, and let $\mathbf{x}_k^{\text{pos}}\!\in\!\mathbb{R}^3$ be the position component of $\mathbf{x}_k$. The instantaneous smooth goal and ocean current directions are defined as
\begin{align}
\mathbf{e}_k &:= \mathbf{x}_g^{\text{pos}} - \mathbf{x}_k^{\text{pos}}, &
\mathbf{\hat e}_k &:= \frac{\mathbf{e}_k}{\|\mathbf{e}_k\|_{\varepsilon_e}}, &
\mathbf{\hat c}_k &:= \frac{\mathbf{v}_{c,k}}{\|\mathbf{v}_{c,k}\|_{\varepsilon_c}},
\end{align}
with smoothed norms
\begin{equation}\label{eq:smoothed_norms}
\|\mathbf{e}_k\|_{\varepsilon_e} := \sqrt{\mathbf{e}_k^\top \mathbf{e}_k + \varepsilon_e},\qquad
\|\mathbf{v}_{c,k}\|_{\varepsilon_c} := \sqrt{\mathbf{v}_{c,k}^\top \mathbf{v}_{c,k} + \varepsilon_c},
\end{equation}
for small $\varepsilon_e,\varepsilon_c>0$ to ensure $C^1$ derivatives. The per-stage \emph{"helpfulness"} scalar $s_k\in[0,1]$ combines angular alignment and ocean current strength:
\begin{equation}
s_k 
\;=\;
\underbrace{\tfrac{1}{2}\!\left(1+\mathbf{\hat e}_k^\top \mathbf{\hat c}_k\right)}_{\text{alignment}\in[0,1]}
\;\cdot\;
\underbrace{\tanh\!\Big(\tfrac{\|\mathbf{v}_{c,k}\|_{\varepsilon_c}}{V_{\text{scale}}}\Big)}_{\text{strength}\in[0,1]}.
\label{eq:sk}
\end{equation}
The \emph{useful-current scale} \(V_{\text{scale}}\) (m/s) maps the ocean current magnitude into \([0,1]\) via 
\(\tanh\) with larger \(V_{\text{scale}}\) requires stronger ocean currents before activation. Hence $s_k\!\approx\!1$ only when the horizon current is both \emph{well aligned} with the goal and \emph{strong}. Otherwise $s_k\!\approx\!0$ and the ocean current-exploitation terms remain inactive. Figure \ref{fig:sk} gives a simplistic geometric visualization of $s_k$ derivation.
\subsection{Stage-gated objective - MCS and STF costs}
The finite-horizon objective is
\begin{equation}
J \;=\; \sum_{k=0}^{N-1} \ell_k \;+\; (\mathbf{x}_N-\mathbf{x}_g)^\top Q_f (\mathbf{x}_N-\mathbf{x}_g)
\label{eq:obj}
\end{equation}

\paragraph{Notation and Stability Guaranties}

Using \eqref{eq:smoothed_norms} the projector onto the goal line and the along-track component are defined as
\[
P_{\parallel,k} := \mathbf{\hat e}_k \mathbf{\hat e}_k^\top \in \mathbb{R}^{3\times 3},\qquad
\mathbf{e}_{\parallel,k} := P_{\parallel,k}\, \mathbf{e}_k \in \mathbb{R}^3 .
\]
In the proposed formulation, only the along–track error $\mathbf{e}_{\parallel,k}$ is subject to relaxation under favorable ocean currents, whereas the cross–track error $\mathbf{e}_{\perp,k}$ remains fully penalized to ensure accurate performance.

We write $\|\mathbf{z}\|_{Q}^2 := \mathbf{z}^\top Q \mathbf{z}$ and set $Q_{\text{pos}} := Q_{1:3,1:3}$. Let $\mathbf{u}_{k,\text{lin}}:=(u_x,u_y,u_z)\in\mathbb{R}^3$, $R_{\text{lin}}:=\mathrm{diag}(R_{11},R_{22},R_{33})$, and define the position velocity $\mathbf{\dot x}_k^{\text{pos}} = \big(f(\mathbf{x}_k,\mathbf{u}_k,\mathbf{v}_{c,k})\big)_{1:3}$. The indicator $\mathbf{1}_{k>0}$ equals $1$ if $k>0$ and $0$ otherwise.

The term $(\mathbf{x}_N-\mathbf{x}_g)^\top Q_f (\mathbf{x}_N-\mathbf{x}_g)$ is the \emph{terminal cost}, used to guarantee stability of the predictive controller, following the stability axioms of \cite{MAYNE2000789}. It is in general formulated by solving the Discrete Algebraic Riccati Equation (DARE) via linearization of the controlled system at the equilibrium.

\paragraph{Stage cost.}
With the "helpfulness" gate $s_k$ from \eqref{eq:sk}, the combined stage cost is

\begin{align}
\ell_k \;=\;&
\underbrace{(\mathbf{x}_k-\mathbf{x}_g)^\top Q (\mathbf{x}_k-\mathbf{x}_g)}_{\text{state tracking}}
\;+\;
\underbrace{\mathbf{u}_k^\top R \mathbf{u}_k}_{\text{effort}}
\label{eq:lk_base1}
\\[-2pt]
&\;+\;
\underbrace{\mathbf{1}_{k>0}(\mathbf{u}_k-\mathbf{u}_{k-1})^\top R_s (\mathbf{u}_k-\mathbf{u}_{k-1})}_{\text{slew}}
\label{eq:lk-base2}
\\[-2pt]
&\underbrace{-\, s_k\,\lambda_{\text{relax}} \;\| \mathbf{e}_{\parallel,k} \|_{Q_{\text{pos}}}^2
\;-\; s_k\,w_{\text{reb}}\; \frac{E_{\text{ref}}}{E_{\text{ref}} + \|\mathbf{u}_{k,\text{lin}}\|_{R_{\text{lin}}}^2}}_{\textbf{Monotone Cost Shaping (MCS)}}
\label{eq:lk-mcs}
\\[-2pt]
&\underbrace{+\; s_k\,\kappa_{\text{eff}}\, \mathbf{u}_k^\top R \mathbf{u}_k
\;+\; s_k\,w_{\text{glide}}\, \big\| \mathbf{\dot x}^{\text{pos}}_k - \mathbf{v}_{c,k} \big\|^2}_{\textbf{Speed-to-Fly (STF)}}.
\label{eq:lk-stf}
\end{align}

The baseline terms \eqref{eq:lk_base1}, \eqref{eq:lk-base2} ensure goal tracking, bounded effort, and smooth actuation. These form the baseline MPC. When horizon currents are aligned and strong ($s_k>0$), \emph{MCS} \eqref{eq:lk-mcs} reduces the penalty \emph{along the goal line} and gives a bounded rebate for small translational thrust, i.e., being non-increasing relative to the baseline objective, while \emph{STF} \eqref{eq:lk-stf} raises the marginal price of thrust and encourages $\mathbf{\dot x}^{\text{pos}}_k \approx \mathbf{v}_{c,k}$ i.e., "gliding". These two terms bias the optimization toward solutions where the ocean current does useful work. If ocean currents are not helpful ($s_k\approx0$), the controller reverts to the baseline behavior.

The tuning parameters govern when and how strongly the controller exploits ocean dynamics, i.e., which ocean currents are considered favorable.
The \emph{relaxation cap} \(\lambda_{\text{relax}}\in[0,1]\) is the maximal fraction by which the along-track position penalty 
\(\|\mathbf{e}_{\parallel,k}\|_{Q_{\text{pos}}}^2\) is reduced when ocean currents are helpful. 
The \emph{rebate weight} \(w_{\text{reb}}\ge 0\) scales a bounded "energy rebate"
encouraging tiny translational thrust; the \emph{rebate knee} \(E_{\text{ref}}>0\) sets where the rebate halves (at \(\|\mathbf{u}_{k,\text{lin}}\|_{R_{\text{lin}}}^2=E_{\text{ref}}\)). 
The \emph{effort premium} \(\kappa_{\text{eff}}\ge 0\) increases the marginal cost of thrust via \(s_k\,\kappa_{\text{eff}}\,\mathbf{u}_k^\top R \mathbf{u}_k\), discouraging propulsion when the flow can carry the vehicle. 
The \emph{glide-match weight} \(w_{\text{glide}}\ge 0\) scales \(s_k\,w_{\text{glide}}\,\|\mathbf{\dot x}_k^{\text{pos}}-\mathbf{v}_{c,k}\|^2\), biasing the ground velocity toward the ocean current. 
Based on the Copernicus Marine Service data, for the Mediterranean, we found \(V_{\text{scale}}=0.05\!-\!0.3\) m/s, \(\lambda_{\text{relax}}=0.7\!-\!0.9\), \(w_{\text{reb}}=0.3\!-\!0.8\), \(E_{\text{ref}}=20\!-\!40\), \(\kappa_{\text{eff}}=1.5\!-\!3.0\), and \(w_{\text{glide}}=0.15\!-\!0.35\) to be robust.

\subsection{Constraints and optimization}
We enforce feasibility constraints like linear/angular velocity limits, and input bounds and solve the nonlinear program with IPOPT at each control step using Multiple Shooting. The ocean current data $(v_{c,0},\dots,v_{c,N-1})$ are updated from~\eqref{eq:vc-sampling} at every iteration, and the solution is warm-started by shifting the previous optimal trajectories. All gates and normalizations use $C^1$ smoothing (e.g., $\|{\cdot}\|_{\varepsilon}$ and $\tanh$), which improves derivative quality for gradient-based NLP solvers. Constraints are summarized in Table \ref{tab:constraints}.

\begin{table}[h!]
\centering
\caption{Constraints used in the MPC formulation}
\begin{tabular}{ll}
\hline
\textbf{Quantity} & \textbf{Constraint} \\
\hline
Roll angle & $|\phi| \leq 1.2 \,\text{rad}$ \\
Pitch angle & $|\theta| \leq 1.2 \,\text{rad}$ \\
Linear velocity & $|u|, |v|, |w| \leq 1.5 \,\text{m/s}$ \\
Angular velocity & $|p|, |q|, |r| \leq 1.5 \,\text{rad/s}$ \\
Surge force & $|\tau_x| \leq 127.26 \,\text{N}$ \\
Sway force & $|\tau_y| \leq 127.26 \,\text{N}$ \\
Heave force & $-80 \leq \tau_z \leq 100 \,\text{N}$ \\
Yaw moment & $|\tau_\psi| \leq 30.78 \,\text{Nm}$ \\
\hline
\end{tabular}
\label{tab:constraints}
\end{table}

\section{Thruster Power Consumption}\label{sec:thruster_power_consumption}
\subsection{Thruster Allocation}\label{sec:thruster_allocation}

The controller computes a desired 4-DOF control wrench
\[
\boldsymbol{\tau}_4 = \begin{bmatrix} X & Y & Z & N \end{bmatrix}^\top ,
\]
corresponding to surge, sway, heave, and yaw.  
The vehicle, however, is actuated by six thrusters, as seen in Figure \ref{fig:blue_rov}, with individual forces
\(\mathbf{T} \in \mathbb{R}^6\).
The mapping from thruster forces to controlled wrench is given by the \(K_4 \in \mathbb{R}^{4\times 6}\) allocation matrix
, such that the delivered wrench is
\(\boldsymbol{\tau}_{4,\text{del}} = K_4 \mathbf{T}\).

The allocation is formulated as a quadratic program (QP):
\begin{align}
\min_{\mathbf{T}\in\mathbb{R}^6} \quad &
\big\|K_4 \mathbf{T} - \boldsymbol{\tau}_4 \big\|_2^2
+ \lambda \|\mathbf{T}\|_2^2,
\label{eq:thruster-alloc-qp-obj}\\[4pt]
\text{s.t.} \quad &
\mathbf{T}_{\min} \;\le\; \mathbf{T} \;\le\; \mathbf{T}_{\max},
\label{eq:thruster-alloc-qp-bounds}
\end{align}
where \(\mathbf{T}_{\min}, \mathbf{T}_{\max}\) are the per-thruster limits from hardware,
and \(\lambda > 0\) is a small regularization factor to distribute load and penalize excessive thrust. This QP is convex and solved online using CasADi/Ipopt.
The result is the thruster force vector \(\mathbf{T}^\star\) that best approximates the commanded wrench
while respecting actuator constraints.

\subsection{Thruster Power Modeling}
The BlueROV2 is actutaed with Blue Robotics T200 thrusters. We model them using a bi-directional power–law relation between the electrical input power $P$ (W) and the generated thrust $T$ (N). 
Official performance data provided by the manufacturer at a fixed supply voltage (16V) were used for calibration. 
Forward- and reverse-thrust datasets were separated and fitted independently, using as a parametric model, the relation
\begin{equation}
    T = a \, P^b,
\end{equation}
where $a>0$ and $b>0$ are empirical coefficients. 
Taking the logarithm of both variables yields a linear model in log–log space, which was identified using least-squares regression. 
Separate coefficients $(a_f, b_f)$ and $(a_r, b_r)$ were obtained for forward ($T\geq 0$) and reverse ($T\leq 0$) operation, respectively, resulting in the bi-directional thruster model
\begin{equation}
    T(P) =
    \begin{cases}
        a_f \, P^{b_f}, & T \geq 0, \\[6pt]
       -a_r \, P^{b_r}, & T \leq 0.
    \end{cases}
\end{equation}

\begin{figure}[h] 
    \centering
    \includegraphics[width=0.75\columnwidth]{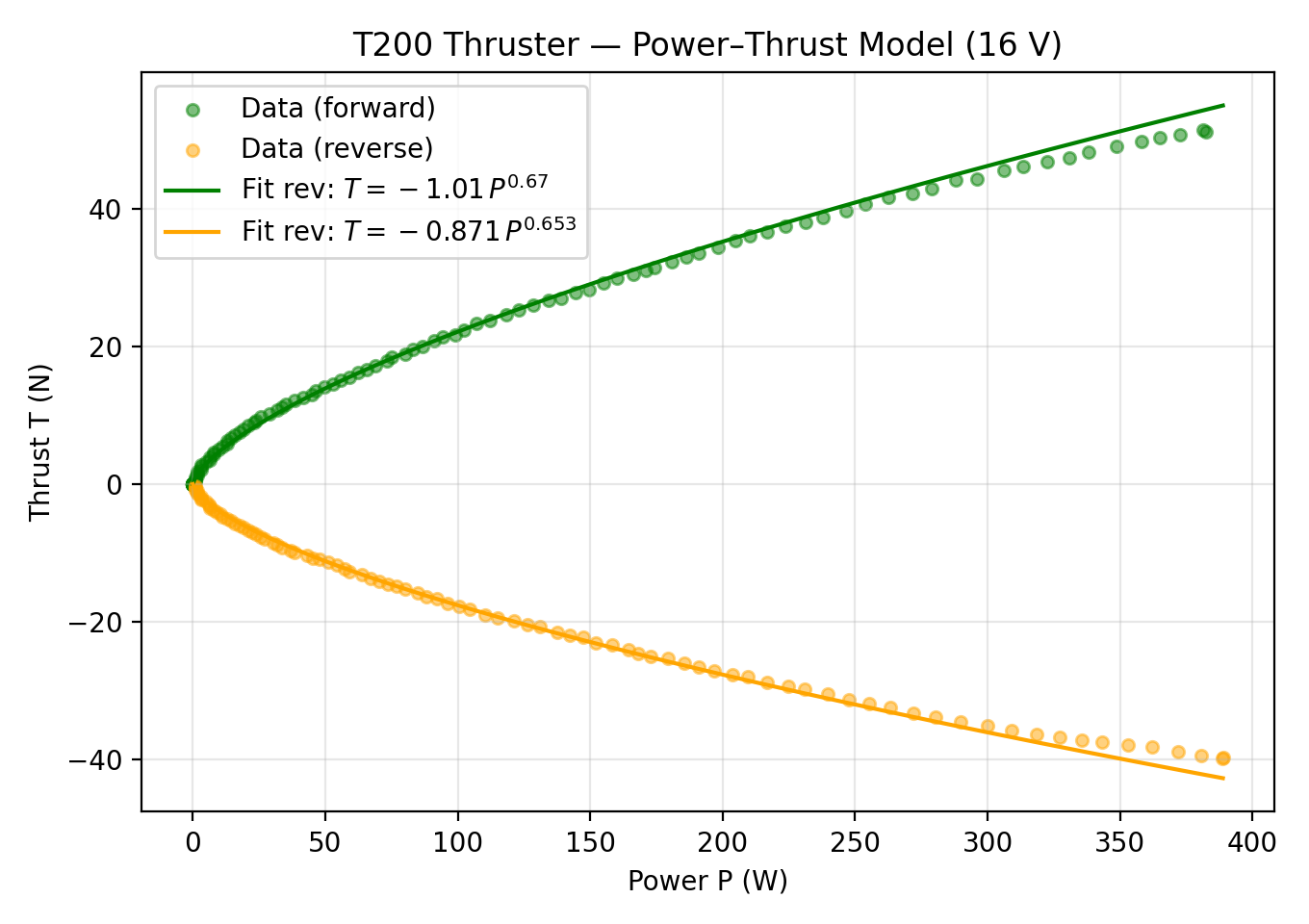} 
    \caption{Thruster power consumption model fitting for 16V.}
    \label{fig:power_fit}
\end{figure}

By solving for $P$, we estimate the required power for a desired thrust. 
This model provides a lightweight yet accurate representation of the T200 thruster suitable for use in real-time control and energy estimation. The fitting curves are depicted in Figure \ref{fig:power_fit}.

\section{Simulation Results}
To highlight energy efficiency, the proposed Current-Harnessing MPC is compared against a baseline MPC that uses only the nominal tracking and effort terms in the stage cost (first three terms), subject to the same dynamic model and constraints. The baseline MPC does not explicitly favor or penalize ocean currents in its objective. However, because the 6-DOF dynamics include relative velocity terms, the baseline controller is still passively influenced by ambient flow and can indirectly benefit from favorable ocean currents, as also noted in \cite{ICRA_2018}. In this work, the effect is even more pronounced since the ocean currents are explicitly propagated throughout the prediction horizon. The proposed Current-Harnessing MPC of course preserves this inherent efficiency by design, while also further exploiting flow conditions through the additional current-aware \emph{MCS} and \emph{STF} terms.

Both controllers are examined across multiple scenarios, with the proposed MPC consistently outperforming the baseline. The controllers only converge to the same performance when no favorable ocean currents are present to be exploited. The comparative results presented here focus on a representative maneuver, which is considered comprising two phases:
\begin{enumerate}
\item A descent from the surface at \((x,y,z)=(40\,\text{m},\,0\,\text{m},\,2\,\text{m})\) to the target depth at \((30\,\text{m},\,40\,\text{m},\,200\,\text{m})\), executed with fixed yaw orientation. Upon approaching \(z=200\,\text{m}\), the vehicle decelerates to near-idle speed.
\item A subsequent horizontal translation at constant depth \(z=200\,\text{m}\) from \((30\,\text{m},\,40\,\text{m},\,200\,\text{m})\) to \((20\,\text{m},\,180\,\text{m},\,200\,\text{m})\).
\end{enumerate}The environment is configured with steady ocean currents, modeled as described in subsection~\ref{sec:ocean_currents}.
The hyperparameter tuning is the same for the two controllers and is described in Table ~\ref{tab:hyperparams}. The weighting matrices are defined as $Q = diag(100,100,100,10,10,10,10,10,10,10,10,10)$, $R = diag(1,1,0.1,0.1)$ and $R_{s} = diag(0.01, 0.01, 0.01, 0.01)$. Both NLPs are solved with high frequency, enabling real-time application of the method.
\begin{table}[h!]
\centering
\caption{Tuning parameters used in the controller design}
\begin{tabular}{ll}
\hline
\textbf{Parameter} & \textbf{Range} \\
\hline
$V_{\text{scale}}$ & $0.05 \text{m/s}$ \\
$\lambda_{\text{relax}}$ & $0.9$ \\
$w_{\text{reb}}$ & $0.8$ \\
$E_{\text{ref}}$ & $40$ \\
$\kappa_{\text{eff}}$ & $3.0$ \\
$w_{\text{glide}}$ & $0.35$ \\
$\Delta t$ & $0.1\text{s}$\\
$N$ & $15$\\
\hline
\end{tabular}
\label{tab:hyperparams}
\end{table}

\subsection{Descent Phase}
\subsubsection{Path and State Trajectories}
\begin{figure}[h!]
    \centering
    \includegraphics[width=0.8\linewidth]{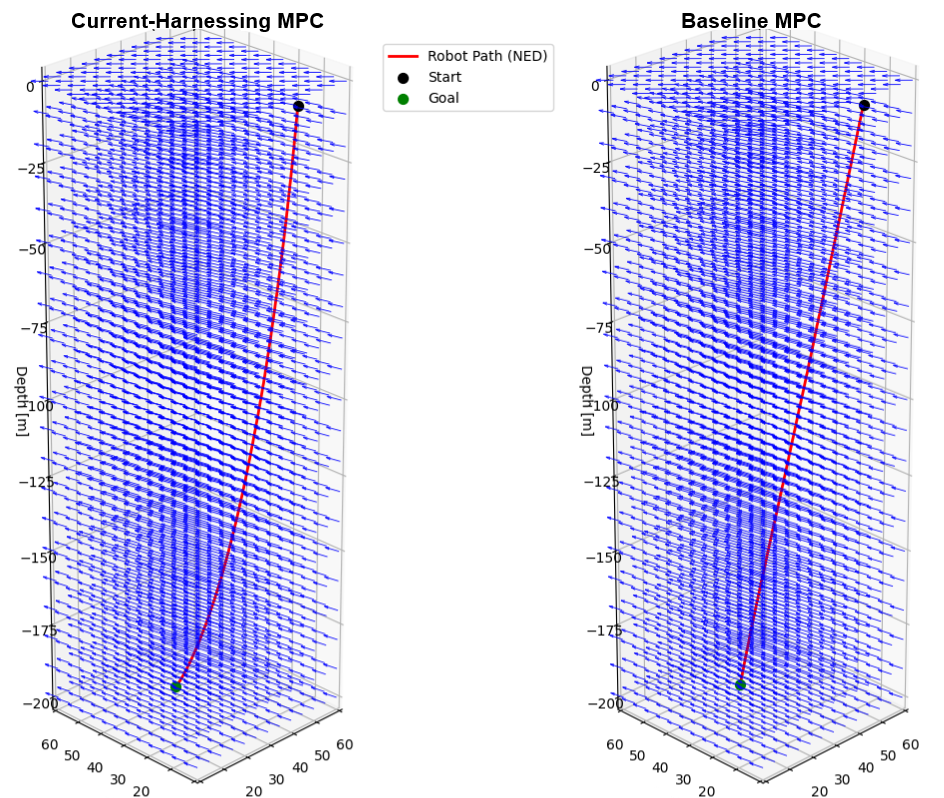}
    \caption{Descent 3D paths.}
    \label{fig:path_compare}
\end{figure}

\begin{figure}[h!]
    \centering
    \includegraphics[width=\linewidth]{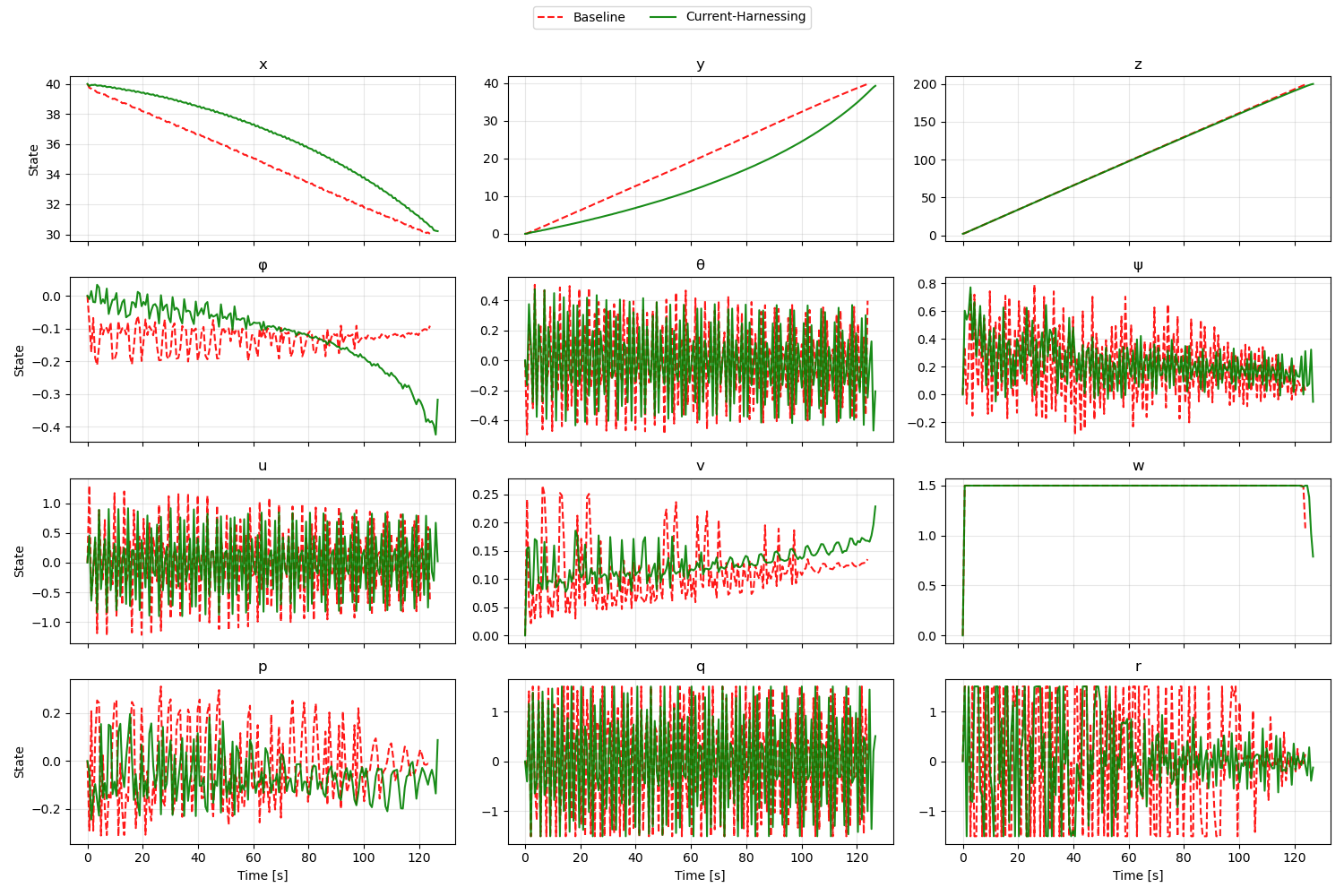}
    \caption{Descent phase state trajectories.}
    \label{fig:trajectories_comarison}
\end{figure}

Figure~\ref{fig:path_compare} shows the 3D path in the NED frame the AUV followed, under both controllers. The small difference curvature is negligible with regard to arrival time, as our proposed method takes $127.2 s$ compared to $124.4 s$ of the Baseline MPC. This difference stems from the Current-Harnessing MPC bending its path to "glide" on the favorable flow. Figure~\ref{fig:trajectories_comarison} depicts a comparison of the corresponding state trajectories. While both controllers achieve depth tracking, the Current-Harnessing MPC exhibits reduced amplitude oscillations.

\subsubsection{Thruster Forces and Allocation}

\begin{figure}[h!]
    \centering
    \includegraphics[width=\linewidth]{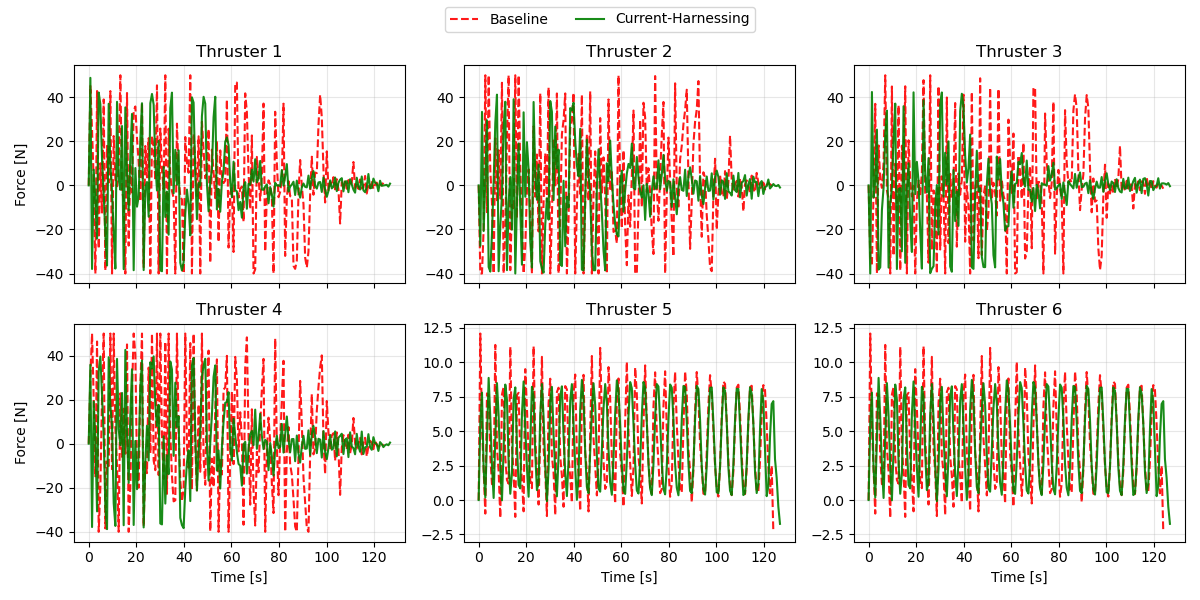}
    \caption{Descent phase thruster forces trajectories.}
    \label{fig:thrusters_compare}
\end{figure}

The generalized forces/torque computed by both controllers are mapped to the six thrusters using the allocation scheme presented in subsection~\ref{sec:thruster_allocation}. 
Figure~\ref{fig:thrusters_compare} shows the resulting thrust trajectory by thruster. 
The baseline MPC uses higher peak forces, while the proposed controller distributes thrust more evenly and stays away from the saturation limits.

\subsubsection{Power Consumption}

\begin{figure}[h!]
    \centering
    \includegraphics[width=\linewidth]{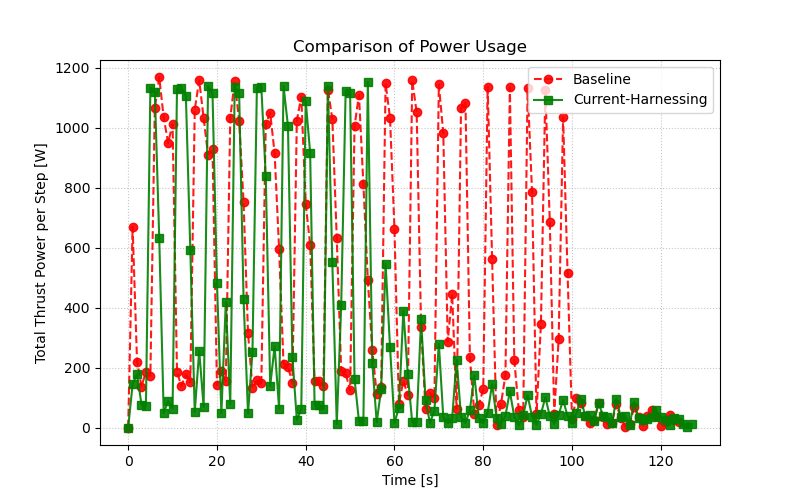}
    \caption{Comparison of instantaneous power consumption during descent phase.}
    \label{fig:power_compare}
\end{figure}

Figure~\ref{fig:power_compare} shows the instantaneous power consumption of the six thrusters for both controllers. 
Table~\ref{tab:energy_compare} summarizes the total mission energy for each controller, along with the mean and maximum energy per time step.
The results confirm that the Current-Harnessing MPC reduces average power demand and achieves the maneuver more efficiently compared to the baseline, with a reduction of $38.4\%$.

\begin{table}[h!]
\centering
\caption{Descent phase energy consumption statistics.}
\begin{tabular}{lccc}
\hline
Controller & Mean [J] & Max [J] & Total [kJ] \\
\hline
Baseline MPC & 46.3 & 142.2 & 57.6 \\
\textbf{Current-Harnessing MPC} & \textbf{27.9} & \textbf{115.4} & \textbf{35.4} \\
\hline
\end{tabular}
\label{tab:energy_compare}
\end{table}

\subsection{Horizontal Phase}
\subsubsection{Path and State Trajectories}
\begin{figure}[h!]
    \centering
    \includegraphics[width=\linewidth]{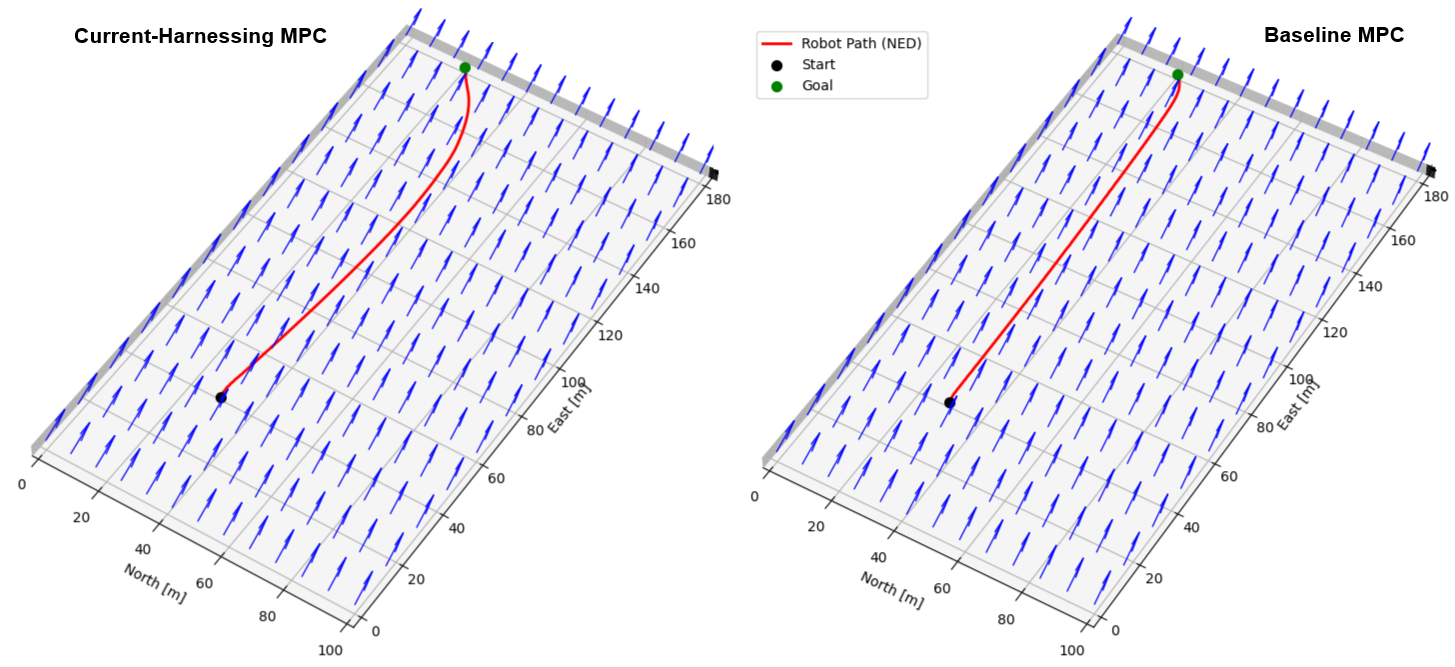}
    \caption{2D paths at 200m depth.}
    \label{fig:path_compare_hor}
\end{figure}

\begin{figure}[h!]
    \centering
    \includegraphics[width=\linewidth]{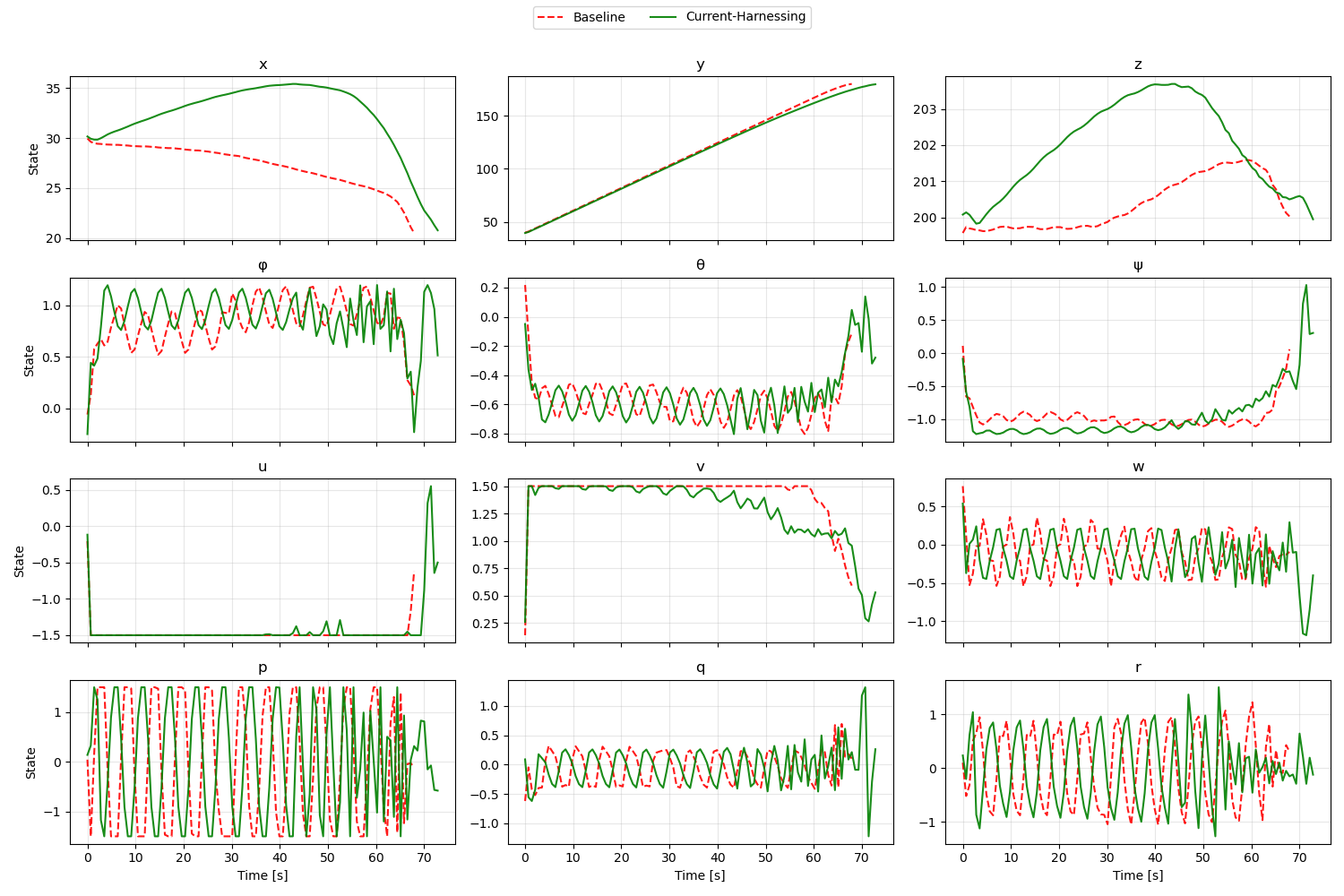}
    \caption{State trajectories of horizontal phase.}
    \label{fig:trajectories_comarison_hor}
\end{figure}

Figure~\ref{fig:path_compare_hor} shows the 2D path in the NED frame the AUV followed during the horizontal movement. In the Current-Harnessing case, the AUV bends towards the favorable ocean current especially at the end of the movement as it slows down and the baseline terms become less dominant. The difference in curvature does not cripple arrival time, as our proposed method takes $73.4 s$ , when the Baseline MPC takes $68.0 s$.
Figure~\ref{fig:trajectories_comarison_hor} depicts the corresponding state trajectories.

\subsubsection{Thruster Forces and Allocation}
Similar to the descent maneuver, the baseline MPC exhibits higher peak force demands, whereas the proposed controller achieves a more balanced thrust distribution and operates away from the saturation limits.

\begin{figure}[h!]
    \centering
    \includegraphics[width=\linewidth]{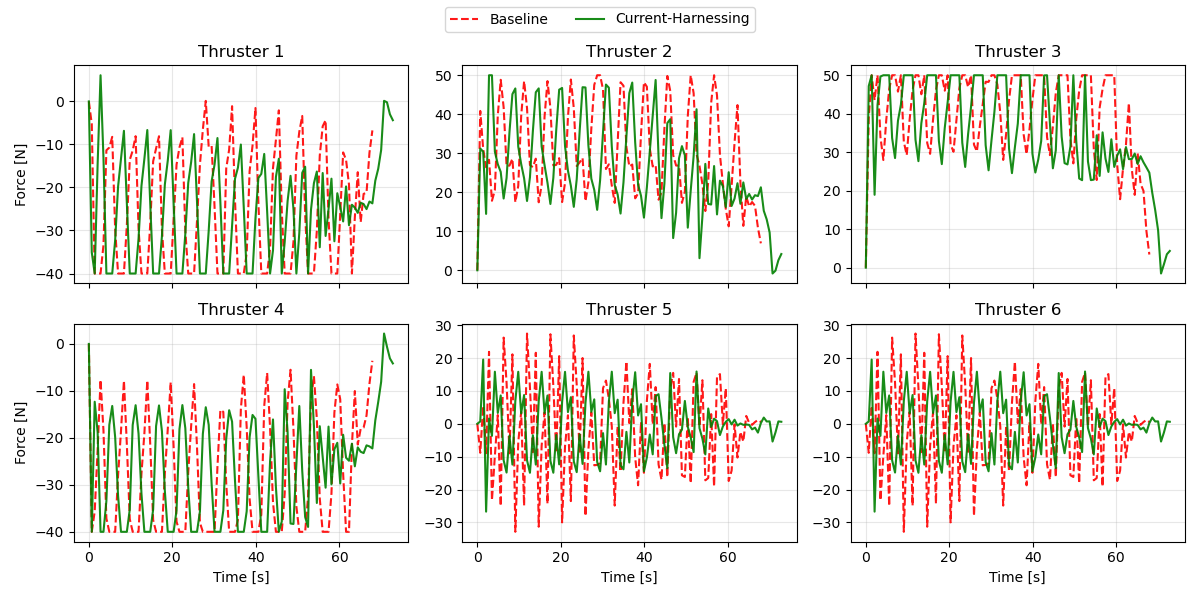}
    \caption{Thrust trajectories of horizontal phase.}
    \label{fig:thrusters_compare_hor}
\end{figure}

\subsubsection{Power Consumption}

\begin{figure}[h!]
    \centering
    \includegraphics[width=\linewidth]{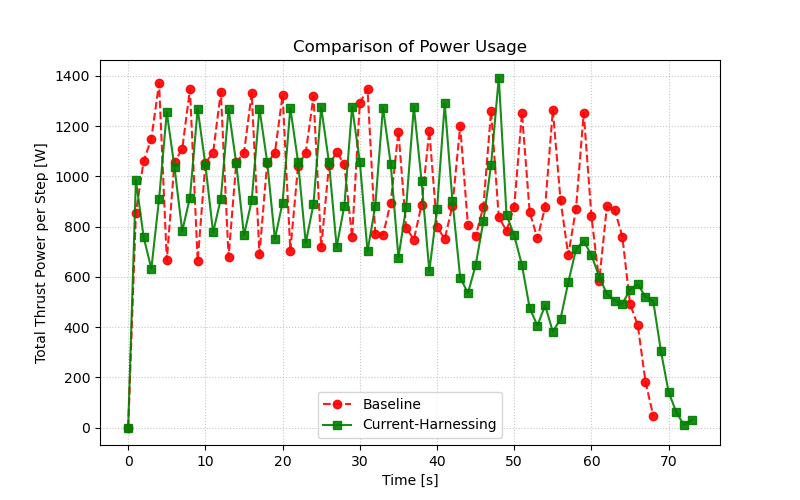}
    \caption{Comparison of instantaneous power consumption during horizontal phase.}
    \label{fig:power_compare_hor}
\end{figure}

Figure~\ref{fig:power_compare_hor} depicts and comperes the instantaneous power consumption of the thrusters.
Table~\ref{tab:energy_compare_hor} presents the total mission energy for each controller, together with the mean and maximum energy per time step.
The results confirm that the Current-Harnessing MPC once again reduces average power demand and achieves the horizontal movement more efficiently compared to the baseline, with a reduction of $12.0\%$.

\begin{table}[h!]
\centering
\caption{Horizontal phase energy consumption statistics.}
\begin{tabular}{lccc}
\hline
Controller & Mean [J] & Max [J] & Total [kJ] \\
\hline
Baseline MPC & 96.2 & 157.0 & 65.5 \\
\textbf{Current-Harnessing MPC} & \textbf{78.7} & \textbf{141.3} & \textbf{57.7} \\
\hline
\end{tabular}
\label{tab:energy_compare_hor}
\end{table}

\section{Conclusions}
We presented Current-Harnessing Stage-Gated MPC, a nonlinear predictive control approach that incorporates stage-gated cost shaping to actively exploit favorable ocean currents without sacrificing baseline performance. The method combines monotone cost shaping with a speed-to-fly penalty, steering trajectories toward energy-efficient behaviors such as near zero water-relative "gliding", while preserving stability and feasibility. Extensive simulations with the BlueROV2 under realistic ocean current conditions show that the proposed controller reduces power consumption relative to a conventional MPC baseline, when ambient flows are favorable, without increasing arrival time or violating vehicle constraints. These findings indicate that the proposed method can significantly extend the endurance of battery-powered AUVs. Future work will investigate experimental sea trials and extend the framework to rapidly varying and uncertain flow forecasts. In addition, the same framework could also be extended by using a fixed corridor direction defined at the beginning of the mission, instead of the instantaneous goal direction used in the proposed method, to further explore potential benefits of a constant reference axis.

\section*{Acknowledgment}
Figure \ref{fig:blue_rov} was edited and improved using Generative AI.

\addtolength{\textheight}{-12cm}   




\bibliographystyle{IEEEtran}
\bibliography{bibliography}

\end{document}